\documentclass{article}
\usepackage{spconf,amsmath,graphicx}
\usepackage{url}
\usepackage{graphicx}
\usepackage{booktabs}
\usepackage{multirow}
\usepackage{color}
\usepackage{multicol}
\usepackage{algorithm}
\usepackage{algorithmic}
\usepackage{amssymb}
\usepackage[table]{xcolor}

\title{Segment Anything Model Meets Image Harmonization}
\name{Haoxing Chen$^{1}$, Yaohui Li$^{2}$, Zhangxuan Gu$^{1}$, Zhuoer Xu$^{1}$,Jun Lan$^{1}$, Huaxiong Li$^{2}$}
 
\address{$^{1}$Ant Group, $^{2}$Nanjing University\\{\tt\small hx.chen@hotmail.com}}

\begin{document}
%
\maketitle
\begin{abstract}
Image harmonization is a crucial technique in image composition that aims to seamlessly match the background by adjusting the foreground of composite images. Current methods adopt either global-level or pixel-level feature matching. Global-level feature matching ignores the proximity prior, treating foreground and background as separate entities. On the other hand, pixel-level feature matching loses contextual information. Therefore, it is necessary to use the information from semantic maps that describe different objects to guide harmonization. In this paper, we propose Semantic-guided Region-aware Instance Normalization (SRIN) that can utilize the semantic segmentation maps output by a pre-trained Segment Anything Model (SAM) to guide the visual consistency learning of foreground and background features. Abundant experiments demonstrate the superiority of our method for image harmonization over state-of-the-art methods.
\end{abstract}
\section{Introduction}
\label{sec:intro}

\begin{figure}[t]
	\centering
	\includegraphics[height=10cm,width=7.8cm]{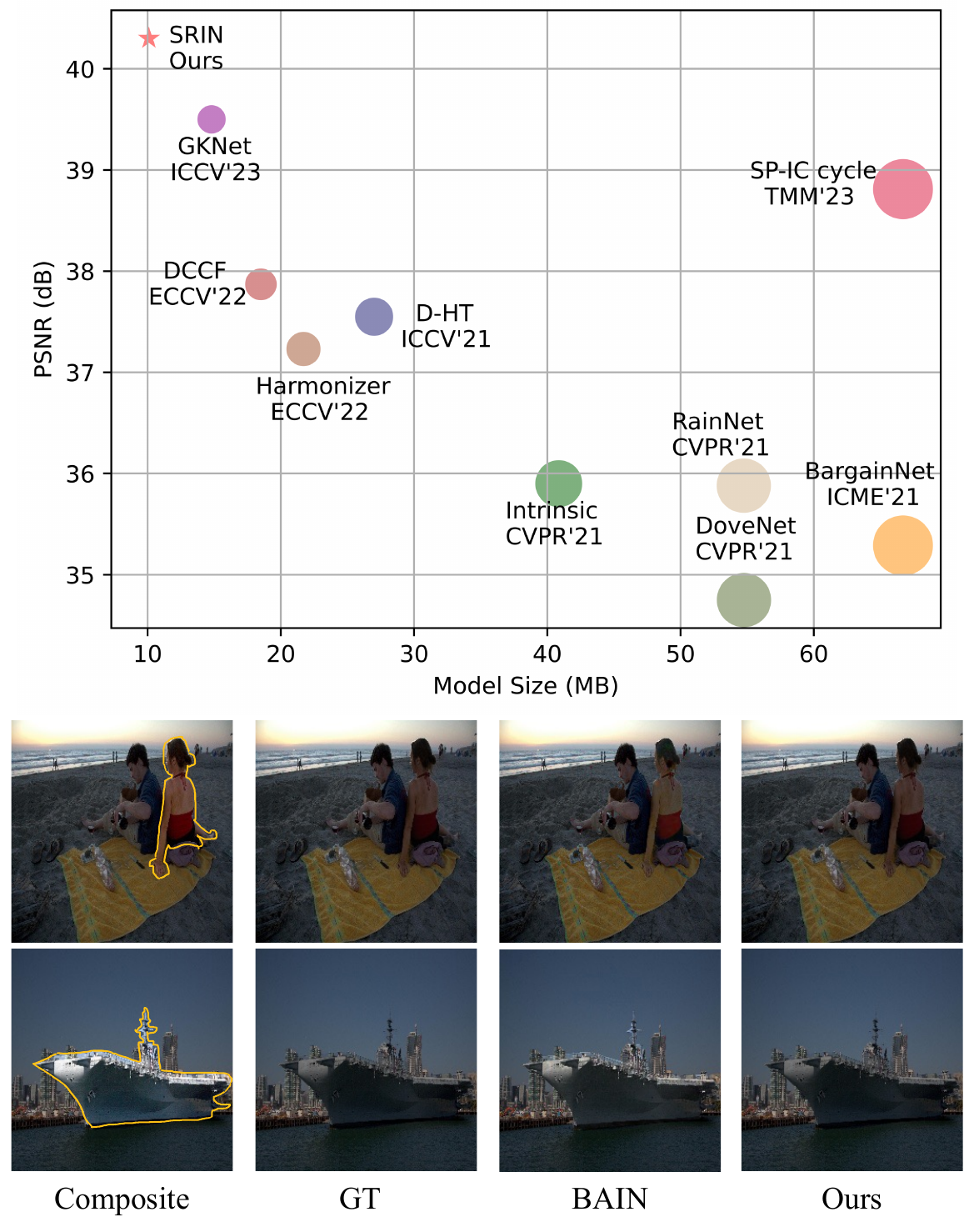}
	\caption{In the top figure, we compare parameter size and performance between our method and other state-of-the-art methods. In the bottom figure, our method produces a more photorealistic harmonized result.}
	\label{fig1}
\end{figure}

Image synthesis~\cite{anydoor,DiffUTE} aims to synthesize foreground objects from one image into another image to produce realistic, visually consistent synthesized images. However, due to differences in lighting, climate, and capture devices, the foreground and background in a synthesized image may have inconsistent appearances.  To address this challenge, image harmonization methods~\cite{Mat2016,RAIN,Bargainnet,MM23_HDNet,DCCF} ensure the compatibility of foreground objects with the background by adjusting their appearance. 

While traditional image harmonization methods~\cite{Mat2016} are mainly based on low-level color and statistical illumination
information, the development of deep learning techniques has made deep image harmonization methods mainstream and achieve better results~\cite{RAIN,Bargainnet,MM23_HDNet,DCCF}. Existing methods can be roughly categorized into two types: local translation and region matching. The former employs an encoder-decoder framework to learn pixel-to-pixel translation~\cite{DIH,SSAM}. However, due to the limitation of shallow neural networks in capturing a limited amount of background, the available background reference for harmonizing the foreground is insufficient. The latter region matching methods treat foreground and background as two distinct style regions. They address the harmonization problem as a matching problem by aligning statistical features or using discriminators on these two regions. Although these methods harmonize images with finer pixel-level features, they completely disregard spatial variations within the two regions. In fact, not all pixels in the background contribute to the harmonization of the foreground, and pixel-level matching may result in the loss of contextual information.

To solve this problem, we rethink the essential semantic priors in image harmonization, i.e., when we paste an object into a background, the color or lighting should be semantically related, and should be referenced based on the semantics of object. Inspired by this observation, we propose a novel semantic-guided region-aware instance normalization (SRIN) that utilizes semantic priors to align the feature distributions of the foreground and background. Specifically, to avoid training an additional segmentation network, we leverage the powerful Segment Anything Model (SAM)~\cite{SAM} to obtain semantic priors. Our method achieves state-of-the-art results on iHarmony4 datasets, which is demonstrated by extensive experiments in both quantitative and qualitative evaluations. As shown in Fig.~\ref{fig1}, the proposed framework demonstrates its efficiency and effectiveness when compared to existing image harmonization models. Notably, our method achieves higher performance while utilizing fewer parameters.

\begin{figure}[t]
	\centering
	\includegraphics[height=8.6cm,width=8.2cm]{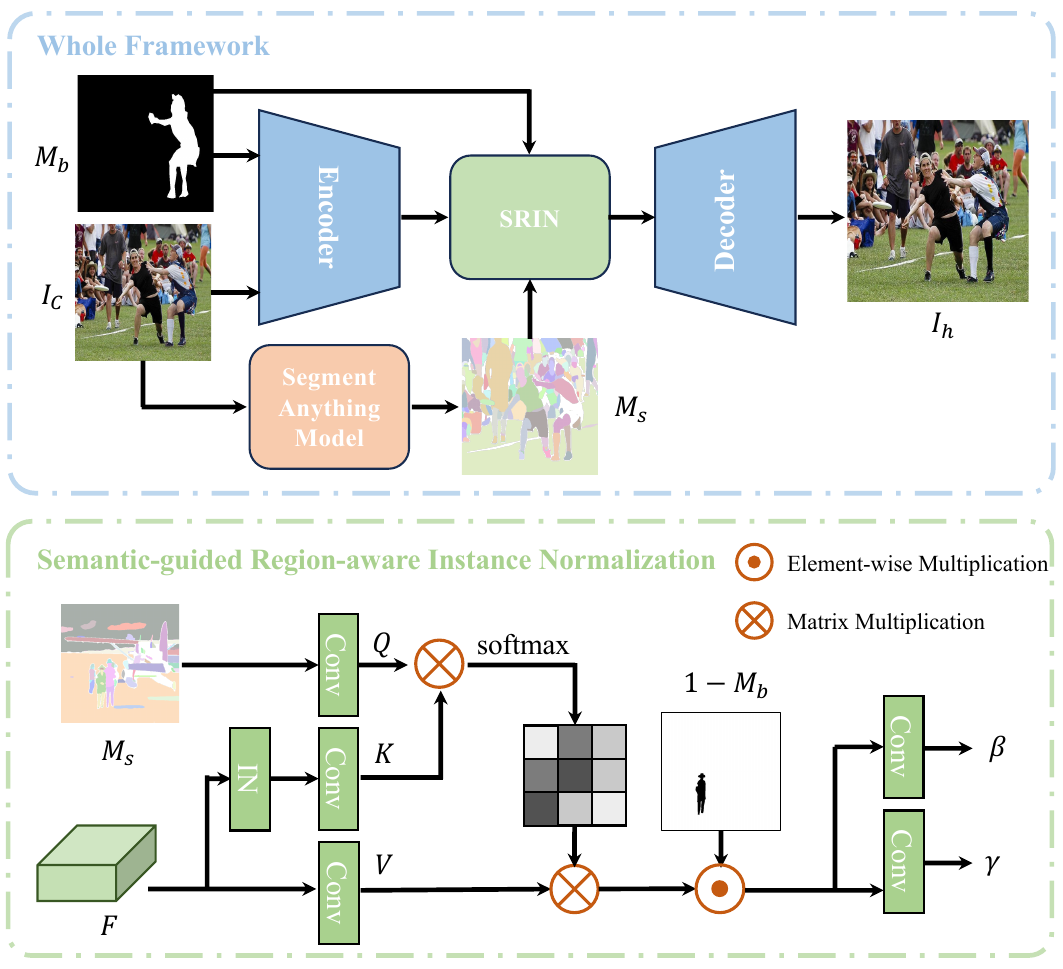}
	\caption{The whole framework of our method and the structure of our proposed Semantic-guided Region-aware Instance Normalization.}
	\label{fig2}
\end{figure}

\section{Method}
\label{sec:method}

\subsection{Problem Formulation}
Image harmonization aims to adjust the appearance of the foreground to make it compatible with the background. An image harmonization task consists of a composite image $I_c\in \mathbb{R}^{3\times H\times W}$ and a binary mask $M_b\in \mathbb{R}^{1\times H\times W}$. Our objective is to learn a harmonization network $G$ that can generate the harmonized image $I_h = G(I_c, M_b) \times M_b + I_c \times \overline{M_b}$. As shown in Fig. 2, our network $G$ is based on U-Net with skip links from the encoder to the decoder. In addition, we propose a semantic-guided region-aware instance normalization (SRIN) inserted between the encoder and decoder. Specifically, SRIN utilizes powerful SAM to obtain semantic segmentation map priors. The L1 loss between harmonized image $I_h$ and ground truth image $I_r$ is used for training $G$ from scratch:
\begin{equation}
    \mathcal{L}_{ih} = ||I_h-I_r||_1.
\end{equation}

\subsection{Semantic-guided Region-aware Instance Normalization}
We illustrate the structure of SRIN in Fig. 2. Given a feature map $F\in \mathbb{R}^{C\times H\times W}$ generated by the encoder, SRIN performs a transformation on the foreground features to align them with the background features, guided by the semantic information provided by the semantic segmentation map $M_s$, where $C$, $H$, $W$ indicate the number of channels, height, and width of $F$, respectively. 

The key of normalization-based methods lies in the generation of the normalization parameters $\gamma$ and $\beta$. To obtain these two parameters, RAIN~\cite{RAIN} estimates the global distribution of the background, while BAIN~\cite{hang2022scs} exploits the weighted background feature distribution by taking into account the pixel-level feature similarity between the foreground and background. However, RAIN only considers the global-level background feature distribution, without considering foreground spatial variations. BAIN only considers the pixel-level feature distribution and alignment, neglecting the semantic information of the objects. To simultaneously incorporate both local details and semantic information, we propose using semantic segmentation maps as priors to guide the alignment of feature distributions. Specifically, we first utilize the SAM to obtain the semantic segmentation map $M_s$ and employ cross-attention between $M_s$ and the feature map $F$ to obtain the semantic-guided attention map $\mathcal{R}$:
\begin{equation}
    Q, K, V = h_q(M_s), h_k(\overline{F}), h_v(F),
\end{equation}
\begin{equation}
    R = {\rm Softmax}(Q^\top \otimes K),
\end{equation}
where $h_q: \mathbb{R}^3 \rightarrow \mathbb{R}^C$, $h_k: \mathbb{R}^C \rightarrow \mathbb{R}^C$, and $h_v: \mathbb{R}^C \rightarrow \mathbb{R}^C$ are $1\times1$ learnable convolution layers, $\overline{F}$ here denotes instance normalized foreground feature map. We then utilize the attention-guided background features to generate the semantic-guided region-aware normalization parameters $\gamma\in \mathbb{R}^{C\times H\times W}$ and $\beta\in \mathbb{R}^{C\times H\times W}$:
\begin{equation}
    \gamma = {\rm ReLU}(g_{\gamma}(V^\top \otimes R \odot M_b)),
\end{equation}
\begin{equation}
    \beta = {\rm ReLU}(g_{\beta}(V^\top \otimes R \odot M_b)),
\end{equation}
where $g_{\gamma}: \mathbb{R}^C \rightarrow \mathbb{R}^C$ and $g_{\beta}: \mathbb{R}^C \rightarrow \mathbb{R}^C$ are $1\times1$ learnable convolution layers. Finally, we get aligned foreground feature $F^n$ at site $(h, w, c)$ by:
\begin{equation}
    F^n_{h, w, c} = \gamma_{h, w, c} \cdot \overline{F}_{h, w, c}+\beta_{h, w, c}.
\end{equation}

\begin{table*}[t]
    \centering
    \caption{Quantitative comparison across four sub-datasets of iHarmony4~\cite{DoveNet}. Top two performance are shown in \textbf{bold}. $\uparrow$ means the higher the better, and $\downarrow$ means the lower the better.}
    \resizebox{\textwidth}{!}{
    \begin{tabular}{cccccccccccc}
    \toprule
       \multirow{2}{*}{Model} & \multirow{2}{*}{Venue} & \multicolumn{2}{c}{HAdobe5k} & \multicolumn{2}{c}{HFlickr} & \multicolumn{2}{c}{HCOCO} & \multicolumn{2}{c}{Hday2night}& \multicolumn{2}{c}{Average} \\ \cline{3-12}&&MSE$\downarrow$&PSNR$\uparrow$&MSE$\downarrow$&PSNR$\uparrow$&MSE$\downarrow$&PSNR$\uparrow$&MSE$\downarrow$&PSNR$\uparrow$&MSE$\downarrow$&PSNR$\uparrow$\\
       \midrule
         DoveNet~\cite{DoveNet}&CVPR'20& 52.32&34.34&133.14&30.21&36.72& 35.83&51.95&35.27&52.33&34.76\\
         RainNet~\cite{RAIN}& CVPR'21 &43.35&36.22 &110.59&31.64 &29.52&37.08&57.40&34.83&40.29& 36.12  \\
        BargainNet~\cite{Bargainnet}&ICME'21& 39.94 &35.34&97.32 & 31.34 & 24.84&37.03& 50.98 &35.67& 37.82 &35.88 \\
        Intrinsic~\cite{GuoZJGZ21} &CVPR'21&43.02&35.20&105.13&31.34&24.92& 37.16 &55.53& 35.96 &38.71&35.90 \\
        D-HT~\cite{IHT}&ICCV'21 &38.53&36.88&74.51&33.13&16.89& 38.76&53.01&37.10&30.30&37.55\\
        Harmonizer~\cite{Harmonizer} &ECCV'22&21.89&37.64&64.81&33.63&17.34& 38.77&33.14& 37.56&24.26&37.84 \\
        S$^2$CRNet-VGG~\cite{liang2021spatial}&ECCV'22&34.91&36.42 &98.73 &32.48 &23.22& 38.48&51.67& 36.81&35.58&37.18 \\
        SCS-Co~\cite{hang2022scs}&CVPR'22&21.01 &38.29 &55.83& 34.22 &13.58 & 39.88 &41.75& 37.83 & 21.33 & 38.75\\
        DCCF~\cite{DCCF}&ECCV'22&23.34 &37.75 &64.77 &33.60&17.07& 38.66&55.76& 37.40&24.65&37.87  \\
        CDTNet~\cite{CDTNet} &CVPR'22&20.62 &38.24&68.61&33.55 &16.25 & 39.15&36.72& 37.95 & 23.75 & 38.23\\
        SP-IC cycle~\cite{cai2023structure}& T-MM'23 & 18.17 & 38.91 & 68.85 &33.88&14.82& 39.73&41.32& 37.90&22.47&38.81  \\
        LEMaRT~\cite{liu2023lemart}& CVPR'23 & 18.80 &39.40 & \textbf{{40.70}} &\textbf{{35.30}} &\textbf{{10.10}}& \textbf{{41.00}} & 42.30& 38.10& \textbf{{16.80}} &\textbf{{39.80}}  \\

        GiftNet~\cite{GiftNEt}&ICCV'23&18.35 &38.76&54.33&34.44 &\textbf{{12.70}}& 39.91&38.28& 37.81 & 19.46 &38.92 \\
        GKNet~\cite{shen2023learning}&ICCV'23&\textbf{{17.84}} &\textbf{{39.97}}&57.58&34.45&12.95&40.32&42.76& 38.47 & 19.90&39.53\\
        \midrule
        \rowcolor{orange!20}Baseline&Ours&24.33 &38.83 & 65.77 & 34.05 & 17.51 & 39.27 &34.08 & 38.15&25.20&38.53\\
        \rowcolor{orange!20}Baseline + BAIN &Ours& 24.72 & 39.06 & 62.56 & 34.37 & 16.11 & 39.58 &\textbf{{32.77}} & \textbf{{38.51}}&24.12&38.83\\
        
        \rowcolor{orange!20}Baseline + SRIN&Ours&\textbf{{14.58}}&\textbf{{41.00}} & \textbf{{50.05}} & \textbf{{35.82}} & 
       14.87 & \textbf{{40.89}} & \textbf{{29.40}} & \textbf{{38.63}} & \textbf{{18.99}} & \textbf{{40.32}} \\
        \bottomrule
    \end{tabular}}
    
    \label{tab:my_label}
\end{table*}

\begin{table*}[t]
    \centering
    \caption{We measure the error of different methods in foreground ratio range based on the whole test set. fMSE indicates the mean square error of the foreground region.}
    \begin{tabular}{cccccccccc}
    \toprule
       \multirow{2}{*}{Model} &\multirow{2}{*}{Venue} & \multicolumn{2}{c}{0\% $\sim$5\%} & \multicolumn{2}{c}{5\% $\sim$15\%} & \multicolumn{2}{c}{15\%$\sim$100\%} &  \multicolumn{2}{c}{Average} \\ \cline{3-10}
       &&MSE$\downarrow$&fMSE$\downarrow$&MSE$\downarrow$&fMSE$\downarrow$&MSE$\downarrow$&fMSE$\downarrow$&MSE$\downarrow$&fMSE$\downarrow$\\
       \midrule
         DoveNet~\cite{DoveNet}&CVPR'20&14.03&591.88 & 44.90&504.42&152.07&505.82&52.36&549.96\\
         RainNet~\cite{RAIN}&CVPR'21&11.66&550.38& 32.05 &378.69&117.41&389.80&40.29&469.60\\
         Intrinsic~\cite{GuoZJGZ21} &CVPR'21&9.97& 441.02&31.51&363.61 &110.22 & 354.84 &38.71& 400.29 \\
         BargainNet~\cite{Bargainnet}&ICME'21& 10.55 & 450.33 & 32.13 & 359.49 & 109.23 & 353.84 & 37.82 & 405.23\\ 
         S$^2$CRNet-VGG~\cite{liang2021spatial}&ECCV'22&6.80&\textbf{{239.94}}& 25.37& 271.70 &103.42 &333.96 &35.58 &274.99 \\
         
        SP-IC cycle~\cite{cai2023structure}& T-MM'23&\textbf{{6.08}} &276.59 & \textbf{{18.27}}&\textbf{{209.56}}& \textbf{{66.44}}& \textbf{{216.37}}&\textbf{{22.47}}&\textbf{{245.75}}  \\
        
         \midrule
         \rowcolor{orange!20}SRIN&Ours&\textbf{{4.58}}&\textbf{{210.49}} & \textbf{{13.98}} & \textbf{{163.29}} & 
       \textbf{{59.18}} & \textbf{{178.58}} & \textbf{{18.99}} & \textbf{{191.08}} \\
        \bottomrule
    \end{tabular}
    \label{tab:my_label}
\end{table*}

\begin{table}[t]
    \centering
    \caption{High-resolution experiments on HAdobe5K.}
    \begin{tabular}{ccccc}
    \toprule
Model&Venue&PSNR$\uparrow$&MSE$\downarrow$&fMSE$\downarrow$\\\midrule
       DoveNet~\cite{DoveNet}&CVPR'20&34.81&51.00& 312.88\\
    Intrinsic~\cite{GuoZJGZ21}&CVPR'21&34.69&56.34&417.33\\
    RainNet~\cite{RAIN}&CVPR'21&36.61&42.56&305.17\\
    CDTNet~\cite{CDTNet}&CVPR'22&\textbf{{38.77}}&\textbf{{21.24}}& \textbf{{152.13}}\\
    INR~\cite{inr}&arxiv'23&38.38&22.68&187.97\\
    \midrule
    \rowcolor{orange!20}SRIN&Ours&\textbf{{40.89}}&\textbf{{14.52}}&\textbf{{126.32}}\\
\bottomrule
    \end{tabular}
    
\end{table}

\noindent
\textbf{Difference with RAIN and BAIN.} RAIN~\cite{RAIN} considers image harmonization as a style transfer problem. However, RAIN only focuses on the style distribution in the background and globally adjusts the second-order statistics of the foreground feature maps to match the statistical characteristics of the background images. On the other hand, BAIN~\cite{hang2022scs} weights the background distribution based on pixel similarity, neglecting the semantic information of the overall objects and losing contextual information. In contrast, our SRIN employs semantic guidance to align the distributions, enabling adaptive adjustment of the feature distribution between each foreground object and similar objects in the background.

\section{Experiment}
\label{sec:experiment}

\subsection{Implementation details}
We evaluate SRIN on the iHarmony4 benchmark~\cite{DoveNet}, following the partition settings of DoveNet \cite{DoveNet}. SRIN is trained from scratch using the Adam optimizer with $\beta_1=0.9$ and $\beta_2=0.999$. The training process lasts for 120 epochs with an initial learning rate of 0.001, which is reduced by a factor of 0.1 at the 100th and 110th epochs. The images are resized to either $256\times256$ or $1024\times1024$, and no data augmentations performed. During the test phase, we evaluate the performance of SRIN using metrics such as MSE, fMSE, and PSNR.

\subsection{Experimental results}
\textbf{Performance on different sub-datasets.} Table 1 lists the quantitative results of previous state-of-the-art methods and our method. From Table 1, we can observe that our method outperforms all of them across all sub-datasets and all metrics. Compared to the most recent method GKNet~\cite{shen2023learning} on iHarmony4 dataset, our method brings 0.91 improvement in terms of MSE, and
0.79 dB improvement in terms of PSNR. 

\begin{figure}[t]
	\centering
	\includegraphics[height=11cm,width=8.6cm]{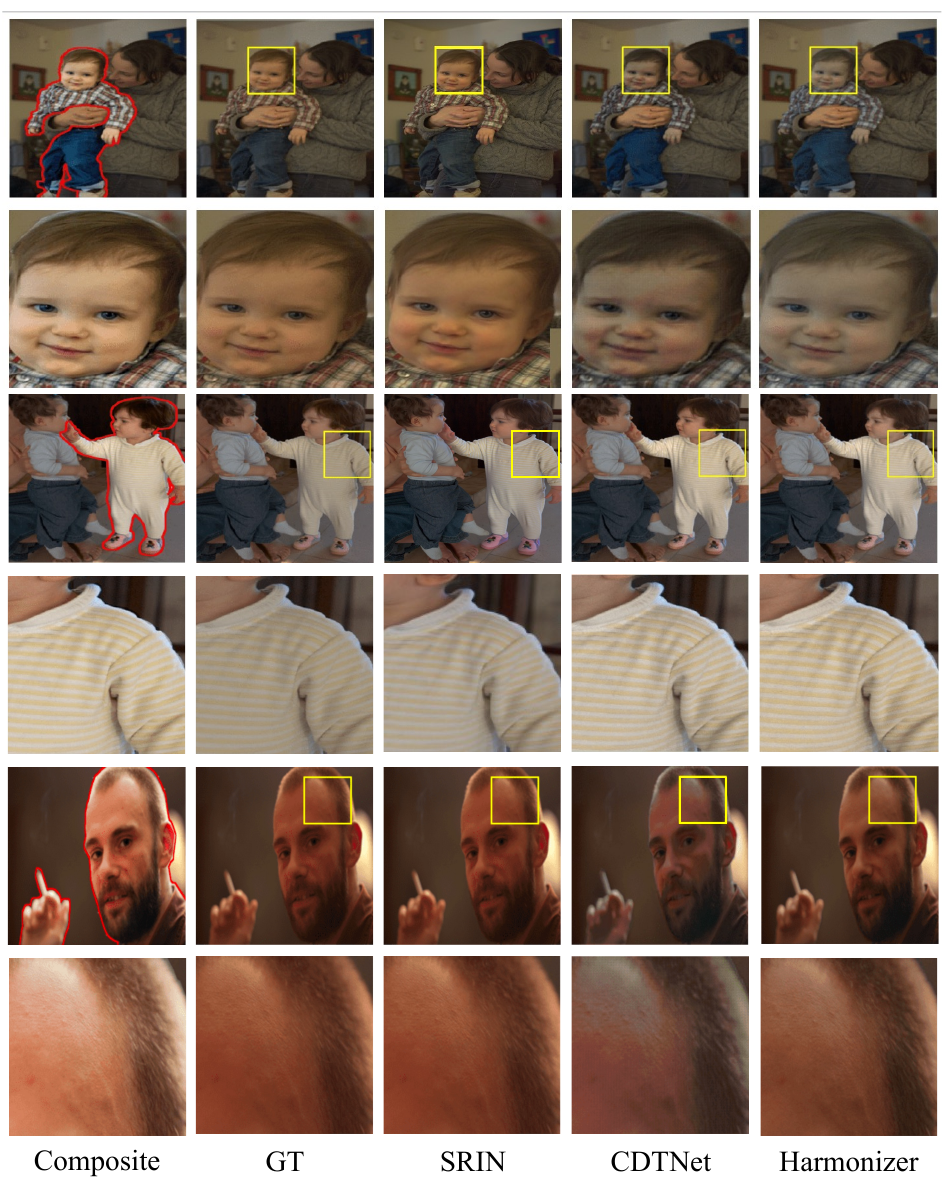}
	\caption{Qualitative comparison on samples from the testing dataset of iHarmony4.}
\end{figure}

\noindent
\textbf{Influence of foreground ratios.} Following~\cite{RAIN}, we examine the influence of different foreground ratios on the harmonization models, i.e., 0\% to 5\%, 5\% to 15\%, 15\% to 100\%, and overall results. The results of all previous methods and our HDNet are given in Table 2. It can be observed that our HDNet achieves the best performance among all approaches. SRIN works well at various foreground scales, thanks to its semantic-guided region dynamics.

\noindent
\textbf{High-resolution results.}
Following~\cite{CDTNet}, we conduct high-resolution experiments. As shown in Table 3, we can see that our method outperforms all of them across all metrics. Compared with the most recent method INR~\cite{inr}, under $1024\times1024$ resolution setting, our method achieves a huge average performance gain of 8.16 in MSE, 61.65 in fMSE, and 2.51 in PSNR. 

\noindent
\textbf{Qualitative comparisons.} We take a closer look at model performance and provide qualitative comparisons with the previous competing methods. From the sample results in Fig. 3, it can be easily observed that our approach integrates the foreground objects into
the background image, achieving much better visual consistency than other methods. The reason why our SRIN can achieve these photorealistic results because it can adaptively learn normalization parameters guided by the semantic segmentation map provided by SAM.

\noindent
\textbf{Effectiveness of SRIN.} To verify the effectiveness of our proposed SRIN, we conducted ablation experiments. Specifically, the baseline is a UNet with the decoder incorporating RAIN~\cite{RAIN}. As the results in Table 1, SRIN achieved better results than RAIN and BAIN on all datasets. Specifically, incorporating SRIN led to a 1.49 dB improvement in PSNR and a 21.3\% reduction in MSE compared to incorporating BAIN. This demonstrates the effectiveness of SRIN and the importance of incorporating semantic priors for harmonization.

\noindent
\textbf{Comparisions on real datasets.} 
We randomly select 100 real composites from the RealHM dataset~\cite{ssh}, making sure there were no duplicate foregrounds or backgrounds. Given each composite image, we can obtain 4 harmonized results including 3 baselines and our method. We invite 50 users to identify the more harmonious one in each pair. Finally 20,000 comparison results are collected, followed by using the Bradley-Terry (B-T) model~\cite{lai2016comparative} to calculate an overall ranking of all methods. As presented in the following Table, our SRIN achieves the highest B-T score.

\begin{table}[t]
\caption{B-T scores comparison on real composite images.}
    \centering
    \begin{tabular}{cccccc}
    \toprule
Model&DCCF&Harmonizer&CDTNet&Ours\\\midrule
B-T Score&0.089&0.247&0.294&\textbf{0.368}\\
\bottomrule
    \end{tabular}
\end{table}

\section{Conclusion}
\label{sec:conclusion}
In this paper, we introduce semantic priors into image harmonization problems and propose a novel Semantic-guided Region-aware Instance Normalization (SRIN) module, which outperforms previous normalization methods by a large margin. SRIN utilizes semantic segmentation maps provided by Segment Anything Model to guide the distribution alignment between foreground and background regions. Experimental results illustrate the superior performance of our SRIN, which has achieved competitive results with other state-of-the-art image harmonization methods.

\newpage
\bibliographystyle{IEEEbib}
\bibliography{refs}

\end{document}